\documentclass[%
 reprint,
 amsmath,amssymb,
 aps,
]{revtex4-2}

\usepackage{graphicx}% Include figure files
\usepackage{float}
\usepackage{dcolumn}% Align table columns on decimal point
\usepackage{bm}% bold math
\usepackage{todonotes}
\usepackage{subcaption} % for subfigures
\usepackage{caption}
\usepackage{placeins}
\usepackage{booktabs}
\usepackage{multirow}
\usepackage{ragged2e}
\usepackage{amsmath}

\begin{document}

\preprint{APS/123-QED}

\title{Neural Diffusion Processes for Physically Interpretable Survival Prediction}
%First-Passage Models with Deep Neural Networks for Survival Prediction}% Force line breaks with \\

% \author{Ann Author}
%  \altaffiliation[Also at ]{Physics Department, XYZ University.}%Lines break automatically or can be forced with \\
%  \email{Second.Author@institution.edu}
% \affiliation{%
%  Authors' institution and/or address\\
%  This line break forced with \textbackslash\textbackslash
% }%
\author{Alessio Cristofoletto}
 \altaffiliation{These authors contributed equally to this work.}
\affiliation{
Department of Computing Sciences, Bocconi University, Milano, Italy
}%

\author{Cesare Rollo}
 \altaffiliation{These authors contributed equally to this work.}
\affiliation{
AI and Computational Biomedicine Unit, University of Torino, Torino, Italy
}%

\author{Giovanni Birolo}
 \altaffiliation{Corresponding author: giovanni.birolo@unito.it}
\affiliation{
AI and Computational Biomedicine Unit, University of Torino, Torino, Italy
}%

\author{Piero Fariselli}
\affiliation{
AI and Computational Biomedicine Unit, University of Torino, Torino, Italy
}%

% \collaboration{MUSO Collaboration}%\noaffiliation

%  \homepage{http://www.Second.institution.edu/~Charlie.Author}
% \affiliation{
%  Second institution and/or address\\
%  This line break forced% with \\
% }%
% \affiliation{
%  Third institution, the second for Charlie Author
% \affiliation{%
%  Authors' institution and/or address\\
%  This line break forced with \textbackslash\textbackslash
% }%

% \collaboration{CLEO Collaboration}%\noaffiliation

\date{\today}% It is always \today, today,

\begin{abstract}

\noindent We introduce \textit{DeepFHT}, a survival-analysis framework that couples deep neural networks with first hitting time (FHT) distributions from stochastic process theory. Time to event is represented as the first passage of a latent diffusion process to an absorbing boundary. A neural network maps input variables to physically meaningful parameters including initial condition, drift, and diffusion, within a chosen FHT process such as Brownian motion, both with drift and driftless. This yields closed-form survival and hazard functions and captures time-varying risk without assuming proportional-hazards. 

\noindent We compare DeepFHT with Cox regression using synthetic and real-world datasets. The method achieves predictive accuracy on par with the state-of-the-art approach, while maintaining a physics-based interpretable parameterization that elucidates the relation between input features and risk. This combination of stochastic process theory and deep learning provides a principled avenue for modeling survival phenomena in complex systems.
%\noindent Survival analysis is an increasingly important tool across fields such as medicine, engineering, and economics, driven by advances in modern data modeling. Recent developments in machine learning and deep learning introduce flexible, high-capacity models that improve predictive performance, but often at the cost of interpretability—an essential requirement in many scientific and applied domains.
%
%\noindent To address this trade-off, we propose DeepFHTs, a novel framework that integrates deep neural networks with the statistical theory of first hitting times (FHT) from stochastic processes. This hybrid approach models the time-to-event as the time of first passage of a latent stochastic process across a barrier; it preserves a physically and probabilistically interpretable structure while leveraging deep learning to predict the parameters of the FHT distribution from input covariates.

%\noindent We evaluate DeepFHTs in the single-risk setting using synthetic and real-world benchmark datasets, and compare its predictive performance to classical survival models. Additionally, we demonstrate how model predictions can be interpreted in the parameter space of the underlying stochastic process. The results show that DeepFHTs achieve competitive accuracy while capturing meaningful and interpretable correlations between features and risk, offering a promising alternative for modeling survival data in complex systems.

\end{abstract}

\keywords{Suggested keywords}%Use showkeys class option if keyword
\maketitle

%\tableofcontents

\section{\label{sec:intro}Introduction}
Survival analysis is central in many applications across medicine, engineering, economics and finance. It concerns \textit{time-to-event modeling}: given a process that can generate an event of interest (e.g., death from disease, failure due to wear), the goal is to estimate the probability that an event occurs at any time $t > 0$ for an individual described by some input variables (or features, or covariates). Unlike standard regression settings, survival data are characterized by censoring, which means that for some instances, the exact event time is not observed (for example, when individuals remain event-free at the end of the study), and only the last recorded follow-up time is available.

%Traditional methods approach this problem by making strong statistical assumptions on the relationship between covariates and risk. The Cox proportional hazards model \cite{cox1972} is the most popular among these approaches and arguably the most well-established method in survival analysis. Proportional hazards assumptions implies the instantaneous risk of an event for two individuals in the dataset differs by a constant factor over time.
%In the original formulation, a simple linear regression is used to model the relationship between covariates and the hazard function, but many other modeling strategies have been adapted to the Cox model since, to introduce non-linearity and improve performance on high-dimensional data \cite{tibshirani1997lasso,ridgeway1999state,faraggi1995neural}.
Traditional approaches to survival modeling rely on strong statistical assumptions linking input variables and risk. The Cox proportional hazards (CoxPH) model \cite{cox1972} remains the most widely used and best established method. The proportional hazards assumption implies that the instantaneous risk of event for two individuals differs by a constant factor over time. The CoxPH model is also linear, making it clear how each single input variable affects the outcome, but at the expense of missing interactions between features. In its original form, this relation is modeled through a linear regression on the features, though many extensions have been developed to relax linearity and improve performance in high-dimensional settings \cite{tibshirani1997lasso,ridgeway1999state,faraggi1995neural}.

Despite its success, Cox regression is limited by the proportional hazards (PH) assumption, which is often unrealistic. Consequently, many alternatives have been proposed that dispense with the PH constraint, ranging from classical statistical formulations \cite{aalen1989,pike1966} to modern methods that incorporate machine learning and deep learning techniques \cite{friedman2001greedy,ishwaran2008rsf,katzman2018deepsurv}.

Among these, \textit{first hitting time} (FHT) models are of particular interest for this work. FHT models assume that event times are distributed as the times of first passage of a latent stochastic process, underlying the event-generating process observed, through an absorbing barrier. This approach overcomes the PH assumption, since FHT distributions for many stochastic processes allow for time-varying hazard ratios between different individuals (e.g., Lévy distribution, Inverse Gaussian distribution). 

%A branch of survival methods aims to model the distribution of first hitting times directly, using machine learning techniques \cite{liu2019hitboost} and deep neural networks \cite{lee2018deephit} to achieve strong predictive performance, but often fails to provide interpretable insights into the nature of the underlying process and how features relate to it, due to the black box nature of most deep learning frameworks. 

A branch of survival methods seeks to model the distribution of event times directly, without imposing a parametric process model. Examples include HitBoost \cite{liu2019hitboost} and DeepHit \cite{lee2018deephit}, which use machine learning and deep neural networks to approximate the event-time density $f(t)$ from data. While these approaches often achieve strong predictive performance, they do so in a purely data-driven manner: the estimated distribution is not tied to parameters of an underlying stochastic process. As a result, the learned models provide limited physical or mechanistic interpretability, and it remains difficult to understand how features influence the latent dynamics that give rise to observed survival outcomes.
Conversely, parametric FHT alternatives \cite{lee2006threshold,debin2023boosting,race2021semiparametric} currently suffer from the opposite drawbacks. These models learn the process parameters on which the distribution of first hitting times depends from data, using simpler regression and machine learning techniques, yielding more physically interpretable results at the expense of model expressiveness.

The present work addresses these complementary limitations by introducing \emph{DeepFHT}, a framework that unites parametric FHT modeling with deep neural networks. In DeepFHT, a neural network maps input features to the parameters of an underlying stochastic process—such as initial condition, drift, and diffusion—that governs the distribution of first hitting times. To our knowledge, this is the first application of modern deep learning to parametric FHT survival models. The hybrid design combines the representational power of neural networks with the mechanistic grounding of FHT processes, yielding closed-form survival and hazard functions while maintaining a direct physical interpretation of parameters. In this way, DeepFHT provides a principled alternative to black-box survival models, providing predictions that are not only accurate but also anchored in a dynamical description that links covariates to the latent process generating events.

We demonstrate the capability of this framework on synthetic and publicly available clinical survival datasets, comparing predictive power against the CoxPH model. DeepFHT achieves competitive performance across clinical benchmarks and clearly outperforms Cox regression on synthetic data designed to probe the model’s capabilities in scenarios with nonlinearities and non-proportional hazards. 

Finally, we assess interpretability by examining the correlations between predicted process parameters, empirical risk, and features. The results show that the similarity between patients outcomes is intuitively encoded as distance in the space of parameters, and that DeepFHT can recover clinically meaningful relationships and has the potential to highlight new correlations in complex, high-dimensional data.

The remainder of the paper is organized as follows. Sections~\ref{sec:survival} and~\ref{sec:fht} provide theoretical background, reviewing the fundamentals of survival analysis and introducing the relevant aspects of stochastic process theory. Section~\ref{sec:methods} presents the proposed models and the experimental setup, while the results and the \textit{physics-based interpretability} of the models are discussed in Section~\ref{sec:results}.

\section{\label{sec:survival}Survival analysis framework}
%\subsection{\label{sec:surv_data}Survival data}
Survival datasets are typically represented as triplets $(\mathbf{x}_i, \delta_i, T_i)$ for $i=1,\dots,n$, where:
\begin{itemize}
    \item $n \in \mathbb{N}$ is the dataset size;
    \item $\mathbf{x}_i \in \mathbb{R}^m$ is a vector of $m$ features;
    \item $\delta_i \in \{0,1\}$ is an indicator that denotes whether the event was observed ($\delta_i = 1$) or the instance was right censored, i.e., end of observation without occurrence of the event ($\delta_i = 0$);
    \item $T_i \in \mathbb{R}_{>0}$ is the time of event occurrence or censoring.
\end{itemize}

The goal of survival analysis is to predict the probability distribution of event times $f(t)$ and some relevant, related functions:
\begin{itemize}
    \item survival function $S(t) = \Pr(T > t)$
    \item cumulative distribution function $F(t) = 1 - S(t)$
    \item hazard function
\begin{equation*}
    h(t) = \lim_{\Delta t \to 0^+} \frac{\Pr(t \leq T < t + \Delta t \mid T \geq t)}{\Delta t}
          = \frac{f(t)}{S(t)} ,
\end{equation*}
\end{itemize}
The survival function describes the probability of surviving beyond time $t$, while the hazard captures the instantaneous risk of event occurrence, given survival up to $t$. Estimation typically focuses on conditional forms $S_{\boldsymbol{\theta}}(t|\mathbf{x})$ and $h_{\boldsymbol{\theta}}(t|\mathbf{x})$, which quantify the dependence of survival dynamics on the features, through a set of parameters $\boldsymbol{\theta}$. If one assumes independent samples and non-informative censoring, the parameters can be determined with a maximum likelihood estimation (MLE) from the data \cite{rollo2025deep}.

A widely adopted semi-parametric approach for modeling how covariates affect the time-to-event distribution is the Cox Proportional Hazards model \cite{cox1972}. It assumes that the hazard rate for an individual with features vector $\mathbf{x}$ is proportional to a baseline hazard function $h_0(t)$ and can be factorized as:
\begin{equation}
h(t \mid \mathbf{x})=h_0(t) \exp \left\{\boldsymbol{\beta}^{\top} \mathbf{x}\right\}, \quad S(t \mid \mathbf{x})=\left[S_0(t)\right]^{\exp \left\{\boldsymbol{\beta}^{\top} \mathbf{x}\right\}}.
\end{equation} 
This assumption implies that the ratio of hazards for any two individuals is constant over time, depending only on their covariates and not on $t$ itself.
Unlike fully parametric methods, CoxPH does not require specifying a functional form for the baseline hazard during the MLE, since the time-dependent components cancel out in the likelihood.

\section{\label{sec:fht}First hitting time models for survival analysis}

A natural probabilistic representation of survival processes is provided by the theory of first hitting times. Consider a one-dimensional continuous-time stochastic process $\{X(t)\}_{t \geq 0}$, with $X(t) \in \mathbb{R}$, described by a probability distribution $p(x,t)$. Let $\Gamma = \{x\leq x_b\} \subset \mathbb{R}$ be a target set and $X(t=0)=x_0 >x_b$ the initial condition. In this setup, $x_b$ can be considered as a barrier between the target set and the initial condition and this is how it will be referred to in the rest of the paper. The first hitting time is then
\begin{equation}
    T = \inf \{ t > 0 : X(t) \in \Gamma \},
\end{equation}
that is, the earliest time at which the process crosses the barrier in $x_b$. When the barrier is absorbing (i.e., killing the process upon transition), the distribution $f(T)$ of first hitting times coincides with the event time distribution in survival analysis. This condition is easily imposed by the Dirichlet condition $p(x_b,t \mid x_0) = 0$. Without losing generalizability, we will always consider positive initial conditions $x_0>0$ and an absorbing barrier in $x_b=0$. This amounts to a generic choice for a process in $X(t) \in \mathbb{R}$.

Since all relevant survival functions can be derived from the distribution of first hitting times $f(T)$, it is crucial to have analytical solutions for this quantity. In this work, we consider two stochastic processes whose theory is well developed and for which an analytical solution exists: the \textit{Brownian motion} and the \textit{arithmetic Brownian motion} (Brownian motion with drift). For a more complete and detailed discussion of FHT theory and models, including but not limited to their application to survival analysis, we refer to \cite{gardiner2009stochastic,lee2006threshold,aalen2008survival}. 

\subsection{\label{sec:brownian}First hitting time and survival distributions for Brownian motion}

Brownian motion is a pure diffusion process described by a probability distribution $p(x,t\mid x_0)$ that evolves according to the Fokker--Planck equation:
\begin{equation}
    \frac{\partial}{\partial t}p(x,t \mid x_0) = D \frac{\partial^2}{\partial x^2}p(x,t \mid x_0),
    \label{eq:fp-bm}
\end{equation}
with constant diffusion coefficient $D>0$ and initial condition $p(x,0 \mid x_0)=\delta(x-x_0)$. Imposing an absorbing barrier at $x_b=0$ corresponds to the Dirichlet boundary condition
\[
    p(0,t \mid x_0)=0, \qquad x_0>0.
\]
The method of images yields the transition probability density (exact solution):
\begin{equation}
\small
    p(x,t \mid x_0) = \frac{1}{\sqrt{4\pi D t}}
    \left[
        \exp\!\left(-\frac{(x-x_0)^2}{4Dt}\right)
       - \exp\!\left(-\frac{(x+x_0)^2}{4Dt}\right)
    \right]
    \label{eq:bm-fht}
\end{equation}
This solution describes a Brownian motion starting at $x_0>0$ with an absorbing barrier at the origin.

From the probability distribution \eqref{eq:bm-fht}, all relevant survival quantities can be derived. We focus specifically on:
\begin{itemize}
    \item \textbf{Survival function}
    \begin{equation}
        S(t) = \int_{0}^{\infty} p(x,t \mid x_0)\,dx 
             = \operatorname{erf}\!\left(\frac{x_0}{\sqrt{4Dt}}\right),
        \label{eq:bm-survival}
    \end{equation}

    \item \textbf{Failure density}
    \begin{equation}
    \label{eq:bm-density}
        f(t) = -\frac{d}{dt}S(t)
             = \frac{x_0}{2\sqrt{\pi}} \,
               \frac{\exp\!\big(-x_0^2/(4Dt)\big)}{(Dt)^{3/2}},
    \end{equation}
    the distribution of first hitting times (i.e., event times).
    
\end{itemize}

The Brownian FHT model is characterized by a failure density that has the form of a Lévy distribution, depending only on process parameters $x_0, D$.

\subsection{\label{sec:arbm}First hitting time and survival distributions for arithmetic Brownian motion}

Adding a constant drift term $\mu$ to Brownian motion yields the \emph{arithmetic Brownian motion}. For this process, the Fokker-Planck equation becomes:
\begin{equation}
    \frac{\partial}{\partial t}p(x,t \mid x_0)
    = -\,\mu\,\frac{\partial}{\partial x}p(x,t \mid x_0)
      + D \frac{\partial^2}{\partial x^2}p(x,t \mid x_0),
    \label{eq:fp-arbm}
\end{equation}
with a constant diffusion coefficient $D>0$. For an initial condition $p(x,0 \mid x_0)=\delta(x-x_0)$, the free space solution is Gaussian with the mean shifted linearly in time:
\begin{equation}
    p(x,t \mid x_0)
    = \frac{1}{\sqrt{4 \pi D t}}
      \exp\!\left(-\frac{(x - x_0 - \mu t)^2}{4 D t}\right).
\end{equation}

When introducing an absorbing barrier at $x_b=0$, it becomes, via the method of images:
\begin{equation}
\begin{aligned}
    p(x,t \mid x_0) &= \frac{1}{\sqrt{4 \pi D t}}
        \Bigg[
        \exp\!\left(-\frac{(x - x_0 - \mu t)^2}{4 D t}\right) \\
    &\qquad
        - \exp\!\left(-\frac{\mu x_0}{D}\right)\,
          \exp\!\left(-\frac{(x + x_0 - \mu t)^2}{4 D t}\right)
        \Bigg],
\end{aligned}
\label{eq:arb-fht}
\end{equation}
valid for $x>0$, $x_0>0$.

From \eqref{eq:arb-fht}, the survival analysis quantities follow:
\begin{itemize}
    \item \textbf{Survival function}
    \begin{equation}
    \begin{aligned}
        S(t) &= \int_{0}^{\infty} p(x,t \mid x_0)\,dx  \\
             &= \Phi\!\left(\frac{x_0+\mu t}{\sqrt{2 D t}}\right)
              - \exp\!\left(-\frac{\mu x_0}{D}\right)\,
                \Phi\!\left(\frac{\mu t - x_0}{\sqrt{2 D t}}\right),
    \end{aligned}
    \label{eq:arb-surv}
    \end{equation}
    where $\Phi(\cdot)$ denotes the standard normal cdf;

    \item \textbf{Failure density}
    \begin{equation}
        f(t) = \frac{x_0}{\sqrt{4 \pi D t^3}}\,
               \exp\!\left(-\frac{(x_0+\mu t)^2}{4 D t}\right),
        \label{eq:arb-density}
    \end{equation}
    which corresponds to an inverse Gaussian distribution.

\end{itemize}

In the limit $\mu \to 0$, these expressions reduce to the driftless Brownian case, with the failure density converging to the Lévy distribution. Given sufficient time, failure is certain only for $\mu<0$. 

\section{\label{sec:methods}Methods}
\subsection{\label{sec:model}DeepFHT}

As outlined in Sec.~\ref{sec:fht}, survival outcomes can be modeled by first hitting time (FHT) distributions with absorbing barriers. The corresponding survival functions and event-time distributions are determined by the choice of underlying process, depending explicitly on its parameters.
\onecolumngrid
\begin{center}	\includegraphics[width=0.9 \linewidth]{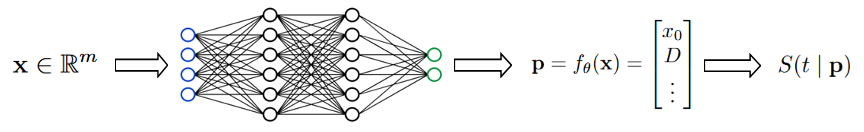}
	\captionof{figure}{Example output of the Lévy FHT model. Individual-specific survival functions are computed from the neural network–predicted parameters.}
	\label{fig:output}
\end{center}
\twocolumngrid
% \begin{figure}[h!]
%     \includegraphics[width=0.5\textwidth]{models.png}
%     \caption{\justifying Example output of the Lévy FHT model. Individual-specific survival functions are computed from the neural network–predicted parameters.}
%     \label{fig:output}
% \end{figure}
To model these functions and distributions from data, we employ a feedforward neural network $f_\theta$ that maps features $\mathbf{x}_i \in \mathbb{R}^m$ to a vector of distributional parameters $\mathbf{p}_i = f_\theta(\mathbf{x}_i) \in \mathbb{R}^d$, i.e. the parameters of the underlying stochastic process. The parameterizations $\mathbf{p}_i$ produced by the network are then used to evaluate survival quantities such as $S(t\mid\mathbf{x}_i) = S(t;\mathbf{p}_i), F(t \mid \mathbf{x}_i) = F(t;\mathbf{p}_i)$, where $S$ and $F$ denote the functional forms of the survival and failure distributions of the chosen FHT model.

The dimension $d$ of the output vector $\mathbf{p}$ depends on the parametrization choice. As anticipated in Section~\ref{sec:fht}, in this paper we will focus on Brownian and arithmetic Brownian motion, identified by their FHT distributions \ref{eq:bm-density},\ref{eq:arb-density}, with the following parameterizations:
\begin{itemize}
    \item \textbf{Lévy distribution:} $\mathbf{p} = \big(x_0, D\big)_{\theta}$,
    \item \textbf{Inverse Gaussian distribution:} $\mathbf{p} = \big(x_0, \mu\big)_{\theta}$ with $\mu<0$, $D=1$,
\end{itemize}
For the inverse--Gaussian model we fix $D=1$ without loss of generality. The FHT law depends on $\mu_T = x_0/|\mu|$ and $\lambda = x_0^{2}/(2D)$, thus $D$ only sets the time scale and can be absorbed by rescaling time (or $\mu$). We therefore parameterize the model by $(x_0,\mu)$ with $\mu<0$.

Notably, both parameterizations above allow for non-proportional hazards, reflecting the flexibility of FHT-based models to capture complex time dependencies in risk. Figure~\ref{fig:output} illustrates a representative model output, where survival functions are computed from the estimated distributional parameters for a set of individuals.
It is also worth noting that this model can easily be extended to any other probability distribution of choice. It is sufficient to specify the functional form and its parameterization and to modify the size of the output layer accordingly.

%In Sec.~\ref{sec:exp_setup} we take advantage of this flexibility to compare results obtained with Lévy and Inverse Gaussian distributions against those obtained with failure density distributions that have no direct interpretation in the first hitting time framework.

\subsubsection*{Physics-based interpretability}
A distinctive feature of DeepFHT is its physics-based interpretability.  Because the model is parametrized by quantities with direct physical meaning (initial condition, drift, and diffusion) each individual is represented as a point in a latent parameter space whose geometry reflects the underlying process dynamics.  The influence of every parameter on event-time distribution is analytically known, allowing regions of higher or lower risk to be distinguished and trajectories of individuals to be compared qualitatively in this space.  In the specific Lévy (\ref{eq:bm-density}) and inverse Gaussian (\ref{eq:arb-density}) cases, the resulting two-dimensional parameterization makes these relationships visually intuitive and enables quantitative notions of similarity (e.g., Euclidean distance) between subjects.

To visualize risk in parameter space, we interpolate event times from the set of uncensored training instances $U=\{\mathbf{p}_i\mid \delta_i=1\}$ via inverse-distance weighting:
\begin{equation}
    T(\mathbf{p})=\frac{\sum_{\mathbf{p}_i\in U} w_i(\mathbf{p})\,T_i}{\sum_{\mathbf{p}_i\in U} w_i(\mathbf{p})},
    \qquad
    w_i(\mathbf{p})=\frac{1}{\sqrt{d(\mathbf{p},\mathbf{p}_i)}}
\end{equation}

Here $d(\cdot,\cdot)$ is a distance in the model parameter space. The resulting $T(\mathbf{p})$ acts as a proxy for risk and supports distance-based similarity analyses.

\subsection{\label{sec:loss}Loss function: Brier loss}
Training is performed by minimizing a custom \textit{Brier loss}, adapted from the Brier score \cite{brier1950verification,graf1999assessment} (see Sec.~\ref{sec:exp_setup}) commonly used to evaluate probabilistic forecasts. The loss is defined as
\begin{equation}
    \begin{aligned}
    \mathcal{L} = \sum_{t \in U} \sum_i \Big[ &I(T_i \leq t \wedge \delta_i=1)\, (S(t|\mathbf{x}_i))^2 \\ &+ I(T_i > t)\, (1-S(t|\mathbf{x}_i))^2 \Big],
\end{aligned}
\label{eq:brier_loss}
\end{equation}
where $U=\{T_j: \delta_j =1\}$ is the set of unique event times where predictions are evaluated. $S(t|\mathbf{x}_i)$ denotes the survival probability predicted for subject $i$, and $I(\cdot)$ is the indicator function.

%While other approaches rely on maximizing the log-likelihood of the survival distribution \cite{faraggi1995neural, lee2018deephit}, we find that in practice the Brier loss yields smoother and more stable optimization dynamics, mitigating vanishing gradients and providing robustness against poor batch composition.

\subsection{\label{sec:exp_setup}Experimental setup}

The experimental evaluation of DeepFHT tests model performance when modeling first hitting time distributions for diffusion processes (Lévy, Inverse Gaussian). As an external benchmark, we also include the Cox proportional hazards model, the standard reference in survival analysis. 

The procedure consists of two phases. First, for each candidate distribution, we perform 5-fold cross-validation on a training set comprising $80\%$ of the available data, with the remaining $20\%$ held out for testing. Each split preserves the original ratio of observed to censored instances. Cross-validation is used to select the optimal architecture and hyperparameters (see Appendix~\ref{sec:appendix_a}), based on average validation performance measured over 100 trials. For hyperparameter search, we use the \textit{Optuna} framework \cite{Akiba2019Optuna}, which implements efficient search strategy for exploring the hyperparameter space. 

After hyperparameter optimization, models are retrained on the full training set and evaluated on the held-out test set. For each performance metric, mean and standard deviation are estimated via bootstrap with resampling: 100 datasets of equal size are drawn with replacement from the test set, each preserving the censoring ratio.  

\subsubsection*{Performance metrics}

\noindent \textbf{Antolini's C-index.} \quad We use the time-dependent concordance index (also referred to as \textit{Antolini's C-index}) \cite{antolini2005time} as the performance metric of choice, for both validation and testing. It is a rank statistic that measures agreement between predicted risks and observed survival times as the probability that, among two comparable individuals, the one experiencing the event earlier is assigned a higher predicted risk (or equivalently a lower survival probability). Antolini's C-index generalizes the original Harrel's C-index \cite{harrell1982evaluating} to models where the predicted risk ranking between individuals is not fixed (like in CoxPH) but changes at different times.
\begin{equation}
\small
\label{eq:antolini}
    c = \frac{\sum_{(i,j)}w(T_i)I(T_i<T_j\land \delta_i = 1 \land S(T_i\mid\mathbf{X}_i) < S(T_i\mid\mathbf{X}_j))}{\sum_{(i,j)}w(T_i)I(T_i<T_j \land \delta_i=1)}.
\end{equation}
where $w(T_i)$ are inverse probability of censoring weights (IPCW) \cite{robins2000correcting}, used to correct for bias introduced by censoring.
This time-dependent version is a censoring-aware equivalent to the ROC-AUC metrics and it is more appropriate for evaluating models whose predictions vary with time.

\vspace{0.5em}

\noindent \textbf{Integrated Brier score.} \quad The Brier score \cite{brier1950verification} measures predictive accuracy as the mean squared difference between predicted survival probabilities and observed outcomes $\delta_i$. Its time-dependent extension \cite{graf1999assessment} evaluates survival probabilities $S(t\mid\mathbf{X}_i)$ at arbitrary times $t$ and uses IPCW. This yields a curve $B(t)$ describing calibration and discrimination across time. To summarize performance over the full observation window, we report the integrated Brier score (IBS),
\begin{equation}
\label{eq:ibs}
    B = \frac{1}{t_{\max}-t_0} \int_{t_0}^{t_{\max}} B(t)\,\mathrm{d}t,
\end{equation}
where lower values indicate better overall predictive accuracy and a Brier score of 0.25 corresponds to a completely uninformative model.
In our experiments, the IBS is used as an additional metric during the testing phase only.

\subsubsection*{Datasets}

We evaluate DeepFHT on publicly available clinical survival datasets, accessed through the \textit{SurvSet} \cite{drysdale2022} repository, as well as on a synthetic dataset designed to test non-proportional hazards and nonlinear relationships between features and risk.  

All datasets undergo identical preprocessing: categorical features are one-hot encoded; missing values are imputed using the mean or median for numerical variables (depending on skewness) and the mode for categorical variables; finally, all features are scaled. 

\vspace{0.5em}

\noindent \textbf{GBSG2:} \quad Data from the German Breast Cancer Study Group 2 \cite{schumacher1994prognostic}, comprising 686 patients with node-positive primary breast cancer. Eight clinical features are available, with $56\%$ censoring.  

\vspace{0.5em}

\noindent \textbf{Framingham:} \quad Derived from the Framingham Heart Study \cite{dawber1951epidemiological}, including 4\,699 participants and 7 cardiovascular health features.  

\vspace{0.5em}

\noindent \textbf{SUPPORT2:} \quad A multi-center observational study of seriously ill hospitalized adults in the United States \cite{knaus1995supportmodel}. The SurvSet version contains 9\,105 records with 35 features; censoring accounts for $32\%$.  

%\vspace{0.5em}
%
%\noindent \textbf{FLChain:} \quad Data from the Mayo Clinic study of free light chain (FLC) testing \cite{dispenzieri2012use}, including 7\,874 patients and 10 features; $72\%$ of cases are censored.  
%
\vspace{0.5em}

\noindent \textbf{NonPH:} \quad A synthetic dataset introduced in \cite{rossi2025beyond}, specifically designed to violate proportional hazards and enforce nonlinear feature–hazard relationships. It contains 10\,000 individuals with 20 normally distributed features. Event times are generated by partitioning the observation period into 16 intervals and assigning individual-specific failure densities:
\[
    p_i = \frac{\exp(16x_i)}{\sum_{j=1}^{16}\exp(16x_j)},
\]
where $x_i$ are features. This construction yields non-proportional hazards by design (see Appendix~\ref{sec:appendix_nonph} for more details).

\section{\label{sec:results}Results}

% C-index across datasets (two-column wide)
\begin{figure*}[t]
  \centering
    \includegraphics[width=2 \columnwidth]{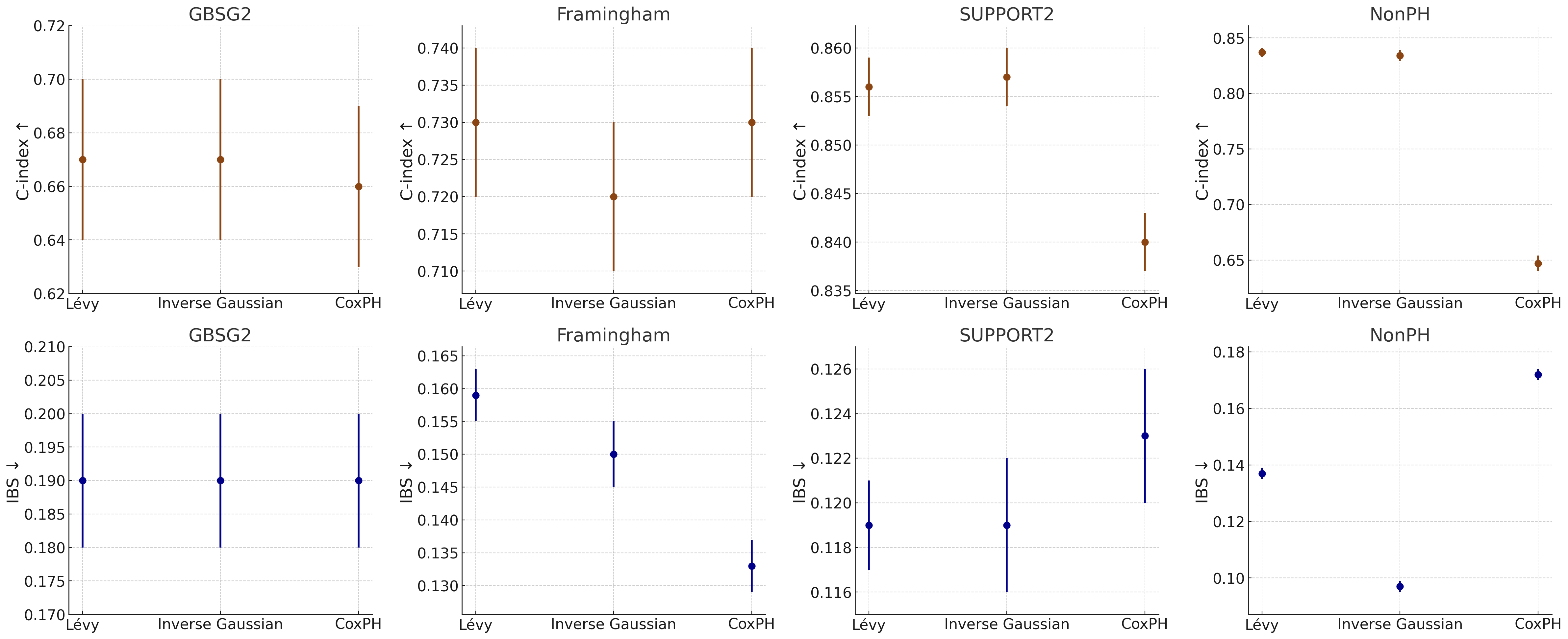}
  \caption{\small \justifying
    Performance across clinical and synthetic datasets. Scatterplots with error bars for C-index (↑, brown, top row) and IBS (↓, dark blue, bottom row) across datasets (GBSG2, Framingham, SUPPORT2, NonPH). 
    For each dataset, three models are shown: Lévy, inverse Gaussian, and CoxPH. 
    Mean values and standard deviations are computed via 100 bootstrap iterations.} 
  \label{fig:performance}
\end{figure*}

\subsection{Model performance}

Predictive test results are summarized in Fig.~\ref{fig:performance}. Across datasets, models based on the more physically interpretable Lévy and inverse Gaussian distributions demonstrates competitive performance, consistently matching or surpassing Cox regression in terms of Antolini's C-index, exhibiting strong robustness and stability at test time.

On the synthetic \textit{NonPH} dataset, DeepFHT significantly outperforms CoxPH with all tested distributions, underscoring the gains achievable in scenarios with nonlinear effects and violations of the proportional hazards assumption.  

Within the clinical datasets, the comparable performance of DeepFHT and CoxPH on GBSG2 and Framingham suggests that in these cases the assumptions of linearity and proportional hazards may not be too restrictive. This points instead to potential limitations in the expressiveness or complexity of the datasets themselves. By contrast, on the higher-dimensional SUPPORT2 dataset, the inverse Gaussian variant achieves a clear improvement over Cox regression, illustrating the advantage of relaxing linear and proportional hazards assumptions in more complex settings.

\subsection{Empirical analysis of parameter-space interpretability}
\label{sec:PX}

%The parametric nature of DeepFHT, grounded in stochastic process theory, provides a natural and physically meaningful basis for interpreting its predictions. As presented in Section ~\ref{sec:model}, for each individual, the model outputs a vector of parameters of the learned underlying diffusion process, effectively representing the subject as a point in a latent parameter space. This space is inherently interpretable: because the influence of each parameter on the process dynamics is known, we are able to distinguish regions of higher or lower risk and to qualitatively compare the trajectories of individuals in this space. Crucially, such analysis can be performed without explicitly computing survival probabilities.

%In our case, the parameterization of Lévy (\ref{eq:bm-density}) and inverse Gaussian (\ref{eq:arb-density}) distributions with DeepFHT yields a two-dimensional parameter space. This representation makes interpretation visually intuitive and, in particular, enables the use of Euclidean distance to assess similarity between instances, i.e., whether they exhibit comparable dynamics and event-time probabilities.
%We show this by visualizing empirical event-time distributions in the parameter space via color gradients and observing that the model consistently assigns previously unseen instances (test set data points) to regions populated by individuals with similar observed event times.

Building upon the interpretability introduced in Section~\ref{sec:model}, we now examine how the learned parameters encode survival dynamics in practice.  Specifically, we visualize event times and clinical features in the two-dimensional parameter spaces of the Lévy and inverse Gaussian models.

\vspace{0.5em}
\noindent \textbf{Event times in parameter space.} \quad %In the parameter spaces of the Lévy and inverse Gaussian models, we visualize the risk associated with each point by interpolating event times from the training data. Only uncensored instances $\mathbf{p}_i \in U = \{\mathbf{p}_i \mid \delta_i = 1\}$ are considered, and interpolation is performed by inverse distance weighting:
%\begin{equation*}
%   T(\mathbf{p}) = \frac{\sum_{\mathbf{p}_i \in U} w_i(\mathbf{p})\, T_i}{\sum_{\mathbf{p}_i \in U} w_i(\mathbf{p})}, 
%    \qquad 
%    w_i(\mathbf{p}) = \frac{1}{\sqrt{d(\mathbf{p}, \mathbf{p}_i)}},
%\end{equation*}
%which assigns to each point $\mathbf{p}$ in the parameter space an estimated event time $T(\mathbf{p})$. These values serve as proxies for patients' risk and allow us to test whether the model implements a similarity notion based on distance in parameter space.% 
Using the IDW interpolation defined in Section~\ref{sec:model}, we map each point in the parameter spaces to an estimated event time $T(\mathbf{p})$, computed from uncensored training instances. These maps serve as risk proxies and allow to test whether the model encodes similarity as proximity in parameter space.
\begin{figure*}[t]
    \centering
    \includegraphics[width=1.04 \columnwidth]{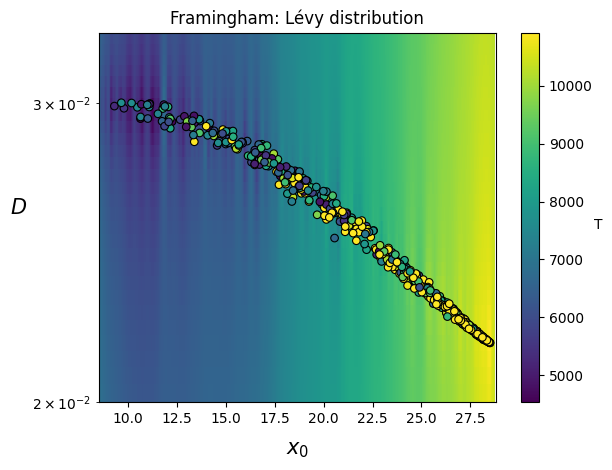}
    \includegraphics[width=1.0\columnwidth]{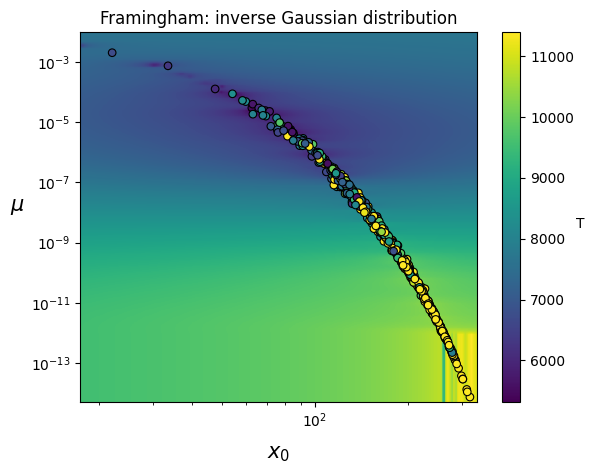}
    \caption{\justifying Event times in the parameter spaces of Deep FHT models. 
    Left: Framingham dataset in the space of Lévy Deep FHT ($\{x_0,D\}$). 
    Right: Framingham dataset in the space of inverse Gaussian Deep FHT ($\{x_0,\mu\}$). 
    Background colors represent interpolated event times $T(\mathbf{p})$ obtained by inverse distance weighting of uncensored training instances. 
    Points correspond to uncensored test patients and individuals surviving beyond the last uncensored time, colored by their observed event times. 
    In both cases, patients with similar event times cluster in contiguous regions, illustrating that the model encodes similarity in terms of process parameters.}
    \label{fig:time_distributions}
\end{figure*}

In Fig.~\ref{fig:time_distributions}, a colormap based on $T(\mathbf{p})$ is superimposed on the parameter space of both Lévy and inverse Gaussian models trained on the Framingham dataset, with uncensored test instances and individuals surviving beyond the last uncensored time represented as points colored by their observed event times. The plots show that previously unseen patients are typically assigned to regions with comparable event times, indicating that individuals with similar risk cluster together in the space of process parameters. This proves to be true across all datasets and models (see Appendix~\ref{sec:appendix_PX}).

We also note that the empirical distribution of event times shown in Fig.~\ref{fig:time_distributions} matches our theoretical knowledge of process dynamics. Patients with longer survival times cluster in regions with favorable initial conditions (i.e., larger distance from the absorbing barrier) and lower diffusion coefficient or drift; processes with these parameter values typically yield longer first-passage times. Conversely, patients experiencing early failure are assigned higher $D$ or $\mu$ and initial conditions closer to the barrier, consistent with shorter first-passage times. 

\vspace{0.5em}

\begin{figure*}[t]
    \centering
    \includegraphics[width=0.7 \columnwidth]{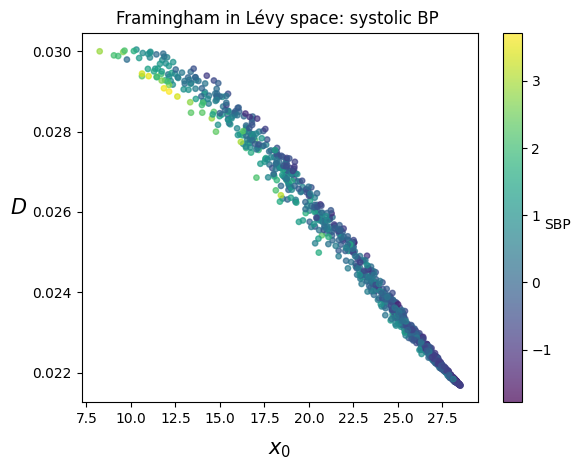}
    \includegraphics[width=0.7\columnwidth]{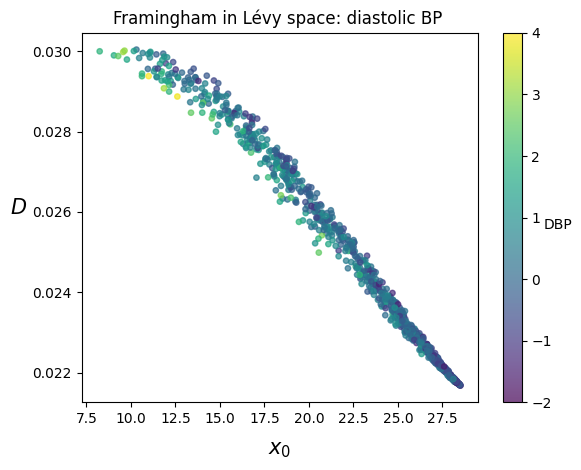}
    \includegraphics[width=0.7\columnwidth]{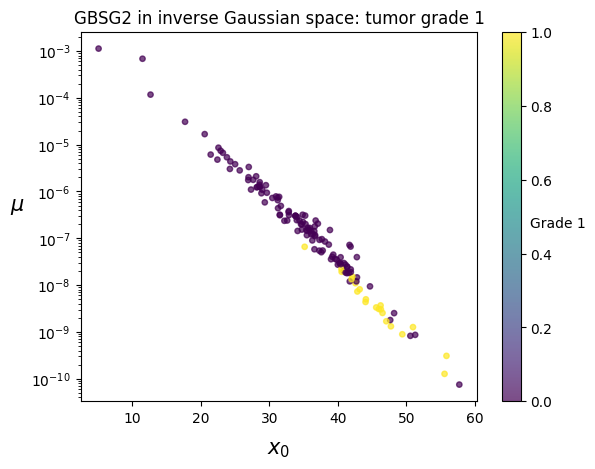}
    \includegraphics[width=0.7\columnwidth]{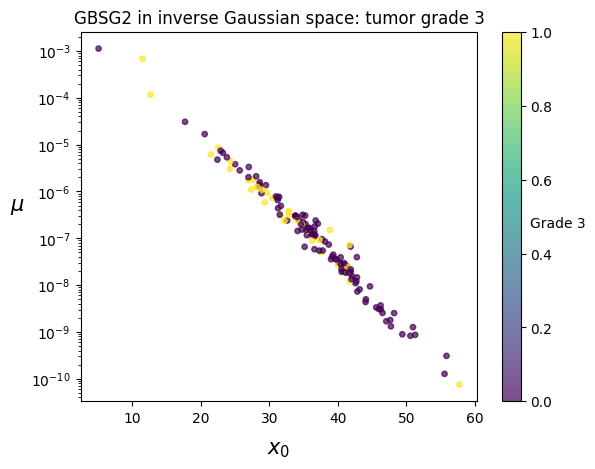}
    \caption{\justifying Feature–parameter relationships in the Lévy and inverse Gaussian DeepFHT models. 
    Top: Framingham dataset with Lévy model, showing the parameter space $\{x_0,D\}$ colored by systolic (left) and diastolic (right) blood pressure.
    Bottom: GBSG2 dataset with inverse Gaussian model, showing the parameter space $\{x_0,\mu\}$ for patients with grade 1 (left) and grade 3 (right) tumors.
    In both cases, clinical risk factors align with model-derived high-risk regions (small $x_0$, large $D$ or $\mu$), supporting the physical interpretability of the parameterization.}
    \label{fig:feature_distributions}
\end{figure*}

\noindent \textbf{Feature values distribution.} \quad We further illustrate physics-based interpretability by examining the relationship between clinical features and the learned distribution parameters. For instance, in the inverse Gaussian model applied to the GBSG2 dataset, patients from the test set are mapped into the parameter space $\{x_0,\mu\}$ and tumor grade is visualized by color gradient (Fig.~\ref{fig:feature_distributions}, bottom row). A clear pattern emerges: patients with low-grade tumors (grade 1) occupy regions with favorable initial conditions and weaker drift (low-risk), while those with high-grade tumors (grade 3) cluster near the absorbing barrier with stronger drift (high-risk). This distribution is consistent with established clinical understanding of tumor aggressiveness and prognosis.

A similar analysis for the Lévy model on the Framingham dataset reveals that individuals with higher systolic blood pressure (\emph{sbp}) and diastolic blood pressure (\emph{dbp}) are more frequently located in the high-risk region of the $\{x_0, D\}$ space (Fig.~\ref{fig:feature_distributions}, top row). For a given diffusion coefficient $D$, these patients tend to have initial conditions $x_0$ closer to the absorbing barrier, linking blood pressure to the only deterministic component of the underlying Brownian motion. This alignment between clinical risk factors and model-derived parameters further supports the physics-based interpretability of DeepFHT.

\vspace{0.5em}

\noindent \textbf{Final remarks.} \quad We conclude this section by noting a phenomenon visible in Figs.~\ref{fig:time_distributions}, ~\ref{fig:feature_distributions}: the apparent collinearity between process parameters (or their logarithms). This effect, observed in both Lévy and inverse Gaussian models, leads to data points concentrating along one-dimensional manifolds in parameter space. As reported in the FHT literature \cite{lee2006threshold}, this behavior can be attributed to the limited complexity of publicly available datasets rather than to intrinsic shortcomings of the models themselves. We reinforce this notion by showing that the parameter space of the synthetic NonPH dataset does not exhibit this effect (see Appendix~\ref{sec:appendix_PX}).

\section{Discussion}

We introduced DeepFHT, a neural survival model that combines the flexibility of deep learning with the physics-based interpretability of first-passage-time distributions. In this work, we focused on diffusion models, for which closed-form survival functions can be derived and whose parameters admit a clear physical interpretation. Importantly, the mapping from input variables to process parameters can be implemented with \textit{arbitrarily complex neural networks}, conferring high modeling flexibility and strong fitting power while \textit{maintaining physics-based interpretability} through the underlying diffusion parameters.

A central advantage of DeepFHT lies in its physics-based interpretability. By mapping individuals to points in parameter space, the model allows us to understand survival in terms of well-characterized process parameters such as initial position, drift, and diffusion. This representation provides intuitive visualizations of survival dynamics and encodes similarity between patient outcomes through distance in parameter space. In this space, DeepFHT reveals clustering of patients by event times and recovers known associations between clinical features and disease outcomes. Such interpretability sidesteps the classical issue of explainability in deep survival models and can be especially valuable in clinical settings where transparency is essential.

Finally, across multiple clinical datasets, DeepFHT achieved predictive performance comparable to or exceeding Cox regression, with clear gains in proof-of-concept scenarios characterized by nonlinear effects and violations of the proportional hazards assumption. Overall, the similar performance of the two approaches suggests limited expressiveness intrinsic to the available datasets. We expect that testing on richer datasets will highlight the value of incorporating deep learning techniques and stochastic models into modern survival analysis.

\FloatBarrier
\bibliography{apssamp}% Produces the bibliography via BibTeX.

\clearpage
\onecolumngrid
\appendix
\section{Model architecture and hyperparameters}
\label{sec:appendix_a}
\textbf{Hyperparameter configurations} \quad Table~\ref{tab:table_1} presents the best hyperparameter configurations obtained via cross-validation, for each dataset and distribution. For all configurations, batch normalization is applied after each hidden layer for improved training stability. Dropout is subsequently applied, if selected during cross-validation, before the activation function.

\noindent The tuned hyperparameters are:
\begin{itemize}
    \item \textbf{Hidden sizes}: Model architecture, as a tuple indicating for each hidden layer.
    \item \textbf{Epochs}: Number of training epochs.
    \item \textbf{BS}: Batch size, computed as \(2^{n}\), where \(n\) is the original tuned exponent.
    \item \textbf{LR}: Learning rate.
    \item \textbf{DO}: Dropout probability.
    \item \textbf{Activation}: Activation function used in hidden layers.
\end{itemize}

\begin{table}[ht]
\centering
\caption{Best hyperparameters per dataset and distribution. Batch normalization is always applied after each hidden layer.}
\scriptsize  % or \scriptsize, or even \footnotesize

\setlength{\tabcolsep}{4pt}  % Reduce horizontal padding between columns

\renewcommand{\arraystretch}{0.95}  % Reduce row height (default is 1.0)
\resizebox{\textwidth}{!}{%
\begin{tabular}{llcccccc}
\toprule
\textbf{Dataset} & \textbf{Model} & \textbf{Hidden Sizes} & \textbf{Activation} & \textbf{DO} & \textbf{Epochs} & \textbf{BS} & \textbf{LR} \\
\midrule 
\multirow{2}{*} {GBSG2}
      & Levy              & 16     & tanh & 0.2 & 150 & 64 & 0.0012 \\
      & InverseGaussian   & 16,16  & elu  & 0.5 & 350 & 32 & 0.0005 \\
\midrule
\multirow{2}{*} {Framingham}
           & Levy         & 16,16  & tanh & 0.5 & 150 & 64 & 0.0011 \\
           & InverseGaussian & 16,16 & elu & 0.2 & 300 & 32 & 0.0003 \\
\midrule
\multirow{2}{*} {SUPPORT2}
         & Levy           & 32,16  & elu  & 0.3 & 300 & 32 & 0.0003 \\
         & InverseGaussian & 32,16 & relu & 0.4 & 200 & 32 & 0.0001 \\
\midrule
%\multirow{2}{*} {FLChain}
%        & Levy            & 16,16  & tanh & 0.1 & 450 & 32 & 0.0006 \\
%        & InverseGaussian & 16     & elu  & 0.5 & 450 & 256 & 0.0030 \\
%\midrule
\multirow{2}{*} {NonPH}
          & Levy          & 16,16  & relu & 0.0 & 450 & 256 & 0.0039 \\
          & InverseGaussian & 16,16 & relu & 0.0 & 200 & 64 & 0.0028 \\
\bottomrule
\end{tabular}%
}
\label{tab:table_1}
\end{table}

\vspace{0.5em}

\noindent \textbf{Cross validation scores} \quad In Table~\ref{tab:table_2} we present validation C-index scores of our model across datasets and distributions. For each pair, we report the mean score across cross-validation folds and the corresponding standard deviation.

\begin{table}[ht]
\centering
\scriptsize
\resizebox{\textwidth}{!}{%
\begin{tabular}{lcccc}
\toprule
\textbf{Model} & \textbf{GBSG2} & \textbf{Framingham} & \textbf{SUPPORT2} & \textbf{NonPH} \\
\midrule
Lévy             & $0.694 \pm 0.017$ & $0.705 \pm 0.012$ & $0.854 \pm 0.004$ & $0.840 \pm 0.004$ \\
Inverse Gaussian & $0.692 \pm 0.024$ & $0.700 \pm 0.013$ & $0.854 \pm 0.005$ & $0.837 \pm 0.010$ \\
\bottomrule
\end{tabular}%
}
\caption{Cross-validation C-indices for Lévy and inverse Gaussian models across datasets.}
\label{tab:table_2}
\end{table}

\clearpage

%\section{NonPH parameter space}
%\label{sec:appendix_b}
%\begin{figure*}[h]
%    \centering
%    \includegraphics[width=0.45 \columnwidth]
%    {appendix_plots/nonph_levy.png}
%    \includegraphics[width=0.45\columnwidth]{appendix_plots/nonph_invgauss.png}
%    \caption{NonPH test set in the parameter spaces of Lévy and inverse Gaussian DeepFHT models. Notice the absence of collinearity between parameters, as discussed in Sec.~\ref{sec:PX}}
%    \label{fig:nonph_params}
%\end{figure*}

\section{\label{sec:appendix_nonph}Synthetic dataset generation}
The synthetic dataset NonPH was constructed to strongly violate Cox Regression assumptions, namely proportional hazards and linearity. Here we describe in detail the procedure used to generate it.

First, we built a matrix of 10,000 observations, each with 20 features \(x_1, \ldots, x_{20}\) randomly sampled from a normal distribution. 
Event times were randomly sampled within the interval \( [0, 10] \), which was discretized into 1,000 equally sized sub-intervals, plus an additional unbounded \( [10,+\infty[ \) interval. 
For each observation, we defined a discrete probability distribution over these intervals as the probability that an event occurs in them according to the a specific survival function parametrized by the features.
A single interval was then sampled, with its lower bound taken as the event time for bounded intervals or as the censoring time for the unbounded interval. This ensured that all events occurred before time 10 and all censoring at time 10.
To enforce consistency, additional censoring was applied so that 25\% of observations were censored. Finally, a random subset of 2,400 observations was retained, preserving the censoring ratio, both to reduce computational cost and to reflect a more realistic dataset size.

The survival distribution used for the event sampling was defined as follows. The time interval \([0,10]\) was partitioned into 16 intervals of equal length where the distribution density is a constant on each interval with value $p_i$ for $i=1,\ldots,16$. The $p_i$ were computed using the soft-max function on the first 16 features (one for each time interval) multiplied by 16: 
$p_i = \frac{\exp(16 x_i)}{\sum^{16}_{j=1} \exp(16 x_j)}$.

The softmax function can be interpreted as a smooth approximation of the argmax operator. When all inputs are scaled by a coefficient \(\beta\), the approximation becomes improves: as \(\beta\) increases, the probability \(p_i\) associated with the largest input \(x_i\) approaches 1, while the others tend toward 0. We set \(\beta = 16\) (coincidentally matching the number of intervals) so that, for most individuals, the event probability is concentrated in a single time interval. With a smaller value such as \(\beta = 1\), the resulting distributions would be flatter and therefore easier to approximate for PH assuming methods. 

\clearpage

\section{\label{sec:appendix_PX}Physics-based explainability: time interpolation plots}
\begin{figure*}[h!]
    \centering
    \includegraphics[width=0.45 \columnwidth]
    {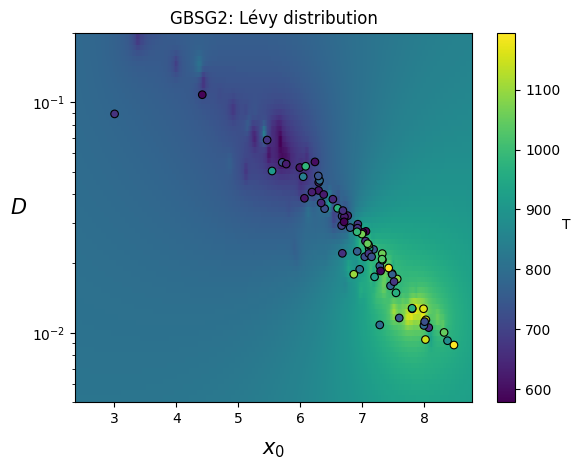}
    \includegraphics[width=0.45\columnwidth]
    {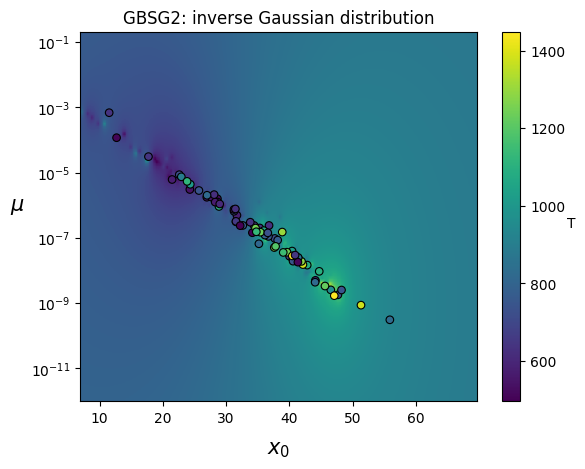}
    \includegraphics[width=0.45\columnwidth]
    {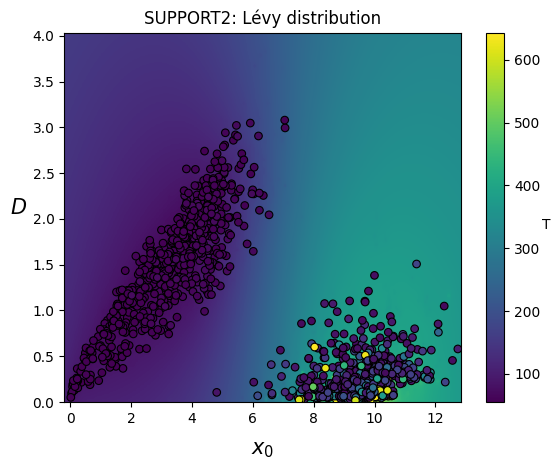}
    \includegraphics[width=0.45\columnwidth]
    {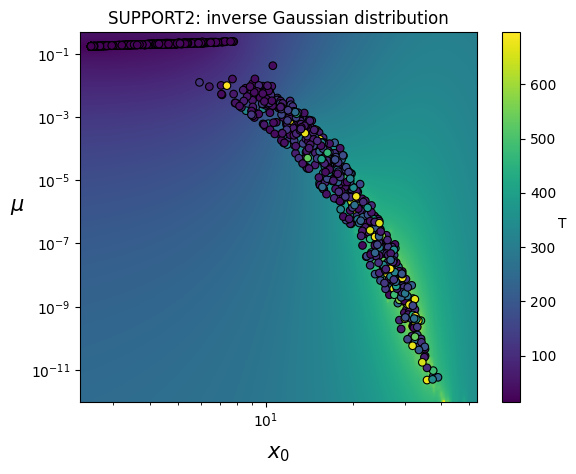}
    \includegraphics[width=0.46\columnwidth]
    {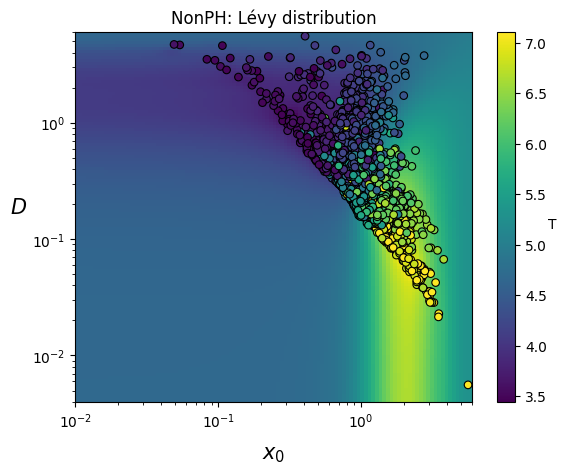}
    \includegraphics[width=0.43\columnwidth]
    {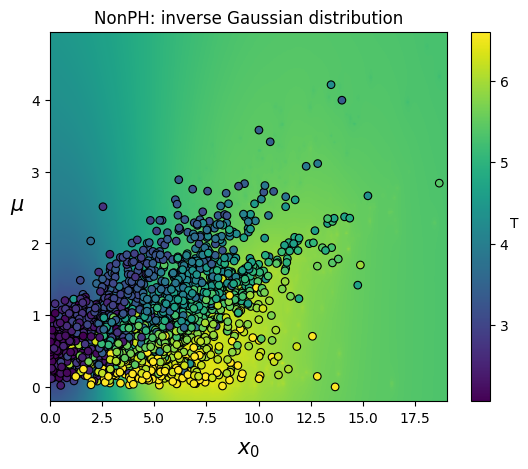}
    \caption{Time interpolation in parameter space across models for GBSG2, SUPPORT2 and NonPH datasets. Notice the absence of collinearity between parameters for the synthetic NonPH datasets, as discussed in Sec.~\ref{sec:PX} }
    \label{fig:appendix_time_interpolation}
\end{figure*}

\end{document}